# A Perspective on Confidence and Its Use in Focusing Attention during Knowledge Acquisition

David Heckerman and Holly Jimison

Medical Computer Science Group
Knowledge Systems Laboratory
Stanford University
Medical School Office Building, Room 215
Stanford, California 94305

## Abstract

We examine a Bayesian approach for accommodating beliefs and preferences that are held with partial confidence. An important notion highlighted by the method is that additional modeling can be valuable when complete confidence is lacking. We develop a meta-decision-analytic approach to balance the benefits of additional modeling with associated costs. We show how the approach can be used during knowledge acquisition to focus the attention of a knowledge engineer or expert on parts of a decision model that deserve additional refinement.

## Introduction

In attempting to automate the reasoning of experts in complex domains, artificial intelligence researchers have explored methods for representing decision problems, including probability and decision theory [1, 2], the Dempster-Shafer theory of belief functions [3, 4], and fuzzy set theory [5]. One argument against decision theory is that it will not accommodate situations in which an expert is unable or unwilling to specify precise beliefs and preferences. Another argument against the theory is that, even if an expert is able to provide assessments, the number of assessments associated with a complex problem is so enormous that the task of model construction becomes unmanageable.

In this paper, we will show that the first argument is incorrect by examining an approach developed within the Bayesian theory for representing beliefs and preferences that are held with partial confidence. More important we will develop the Bayesian perspective on confidence into a methodology for focusing attention during the process of knowledge acquisition. We will see that this approach can increase the efficiency of the model-development task by helping an expert or knowledge engineer to identify portions of a model that are most worthwhile to refine. In developing this approach, we reduce the credibility of the argument that formal models of complex domains are necessarily unmanageable.

First, we examine an approach developed within the Bayesian paradigm for accommodating beliefs and preferences that are held with partial confidence. In doing so, we see that additional modeling often is valuable when assessments are not held with complete confidence. We then develop a method that uses decision theory to balance the benefits of additional modeling with associated costs, and we discuss how the method can be used to focus attention during knowledge acquisition. We next relate our approach to existing methodologies for representing confidence and focusing attention. Finally, we discuss questions raised by our approach and suggest areas for further research.

## A Bayesian perspective on confidence

A method for accommodating beliefs and preferences held with partial confidence has been discussed in the Bayesian literature for over two decades [6, 7]. The basic notion underlying the method is that a person does not feel completely confident in assessing the probability of an event when the probability is influenced by other events with unknown outcomes. That is, a person will be uncomfortable providing a point probability for an event whenever the probability depends on the occurrence or nonoccurrence of other events. This principle applies to preferences held with partial confidence as well.

Given this perspective, the method for accommodating lack of confidence is straightforward. Suppose a person wishes to assess the probability of an event X but is unable or unwilling to provide a precise value. The assessment procedure begins by asking the person to identify other events that have outcomes which influence the probability of X. Once these *influencing* or *conditioning* events are identified, the person typically is able to supply probabilities for the event X conditioned on each of the possible outcomes of the influencing events. If not, additional influencing events are identified until the person is comfortable in providing assessments. Next, the person is asked to supply probabilities for the influencing events. This procedure is applied recursively to any event for which the person is uncomfortable supplying a point value.[1] The probability of X can then be computed from the conditional and unconditional assessments using the laws of probability [8]. This process has been called *extending the conversation* [6]. An analogous procedure exists for assessing preferences held with partial confidence.

To illustrate the approach, suppose we ask a person for his probability that a given football team will win a forthcoming game.[2] We discover that he is uncomfortable supplying a point value because his probability for a win depends on three factors: (1) whether a star player, recently threatened with suspension by league authorities, will be able to play, (2) whether the playing field will be dry or muddy because of rain, and (3) whether the promise of a bonus to the winning players will be confirmed. We will label these events *Sus*, *Field*, and *Bonus*, respectively.

At this point, suppose the person is willing to provide point values for the probability of a win conditioned on each of the possible outcomes of the three conditioning events. A hypothetical set of assessments is shown in the second column of Table 1. Also suppose that the person is able to assess point values for each of the influencing events, say p(Suspension) = 0.6, p(Dry-Field) = 0.7, p(Bonus) = 0.2. Finally, assume that the person believes that the influencing events are independent, so we can compute the probability of each combination of outcomes as shown in the third column of Table 1.

A plot of the third versus the second column of Table 1 is shown in Figure 1. This plot graphically depicts the nature of partial confidence in the person's assessment of p(Win). The quantity p(Win) varies as a function of the outcomes of the influencing events. In general, the wider the "curve" in such a plot, the lower the confidence in the assessment.

Finally, as we mentioned, the laws of probability can be used to compute the person's probability of a win from these assessments. In particular,

```
p(Win) = Σ p(Win | Sus, Field, Bonus) p(Sus, Field, Bonus) = 0.53.
```

where the sum ranges over all possible outcomes of the influencing events.

---

[1] An infinite regress is possible, in theory; however, the focusing method described in this paper offers a practical termination condition.

[2] This example is taken from [7].

| Conditioning events | | | p(Win\|Events) | p(Events) |
|---|---|---|---|---|
| No-sus | Dry-Field | Bonus | 0.7 | (0.4)(0.7)(0.2) = 0.06 |
| No-sus | Dry-Field | No-Bonus | 0.6 | (0.4)(0.7)(0.8) = 0.22 |
| No-sus | Wet-Field | Bonus | 0.6 | (0.4)(0.3)(0.2) = 0.02 |
| No-sus | Wet-Field | No-Bonus | 0.5 | (0.4)(0.3)(0.8) = 0.10 |
| Sus | Dry-Field | Bonus | 0.6 | (0.6)(0.7)(0.2) = 0.08 |
| Sus | Dry-Field | No-Bonus | 0.5 | (0.6)(0.7)(0.8) = 0.34 |
| Sus | Wet-Field | Bonus | 0.5 | (0.6)(0.3)(0.2) = 0.04 |
| Sus | Wet-Field | No-Bonus | 0.4 | (0.6)(0.3)(0.8) = 0.14 |

Table 1: The probability of a win conditioned on relevant events.

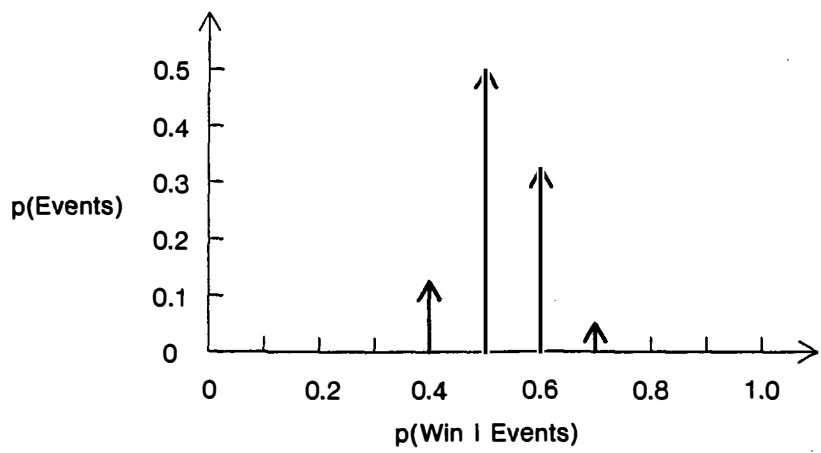

Figure 1: A plot of the assessments in Table 1.

## Benefits of additional knowledge acquisition

The process of identifying partial confidence in assessments and extending the conversation can be quite valuable in a decision context. One benefit is that the process suggests experiments that can be done to update probabilities and utilities before decisions based on these quantities are made. Another benefit is that the process of extending the conversation itself can lead to improved probability and utility assessments. In both cases, the improved assessments can lead to decisions of higher expected value.

To illustrate the first point, suppose our probability assessor of the previous section is offered the chance to gamble on the outcome of the game. If he decides to bet, he will receive $5000 if his team wins, and he will have to pay $5000 if his team loses. The decision tree for this situation is shown in Figure 2. Assuming the person is an expected-value decision maker,[3] he will prefer the alternative of accepting the bet, which has an expected value of $300.

Now, if the decision maker extends the conversation, a potentially valuable option is immediately suggested. In particular, he might be better off if he can delay the decision of whether to bet until just before the game starts, perhaps by buying this option from his fellow gambler. Delaying the bet is potentially valuable because the outcomes of the conditioning events will be revealed before the start of the game. Knowing the outcomes, the decision maker can selectively make or avoid the bet to his benefit. For example, if the star player is suspended, the playing field is muddy, and the bonus offer is denied, the decision maker's probability for a win becomes 0.4. By avoiding the bet in this situation, the decision maker saves $1000 in expected value. Because this conjunction of outcomes for the conditioning

---

[3]This assumption is made to simplify computation. It is by no means necessary.

events occurs with probability 0.14 (see Table 1), his net expected savings due to delaying the decision is $140. More formally, the value of delaying the decision in this case is equal to the *value of perfect information* [9] on the three conditioning variables.

Extending the conversation has suggested a simple delay option which has increased the expected value of the situation. Of course, we can imagine other situations in which more complex experiments might be designed and implemented to update the probabilities and utilities associated with a future decision. In general, the absence of complete confidence in a probability or utility assessment should serve as a flag that additional information gathering might be valuable.

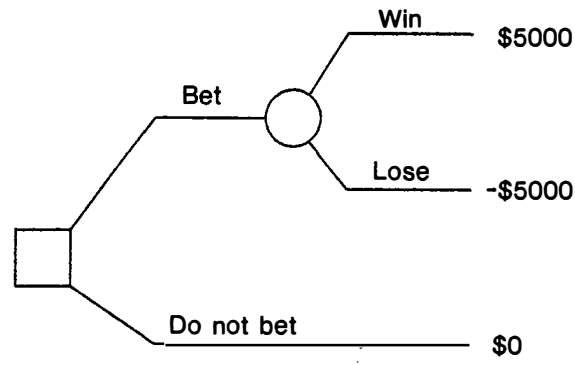

Figure 2: A decision tree for a football bet.

As we mentioned, extending the conversation also is valuable in that the process itself can lead to improved assessments and therefore to decisions of higher expected value. In the situation described, for example, it is unlikely that the individual would give a probability of 0.53 for p(Win) prior to extension. It is far more likely that he would give some coarse value such as 0.4, 0.5, or 0.6. Thus, if our probability assessor does not model his confidence in assessing p(Win), there is a reasonable chance that he will decline the bet, thereby throwing away the opportunity to earn an expected $300. Underlying this argument, is the assumption that the assessments produced by extending the conversation more accurately reflect the beliefs of the decision maker. This assumption is often validated in practice, in the sense that an assessor usually prefers to use the assessments generated from extensions over direct assessments. We will make this assumption throughout the paper.

We have distinguished two classes of benefits for the purposes of this presentation. In practice, however, both benefits typically are realized in the process of extending the conversation. For example, in recognizing partial confidence in the assessment for p(Win) and consequently extending the conversation, our gambler will benefit both by improving his current assessment of p(Win) and by discovering the delay option.

## An approach for focusing attention during knowledge acquisition

The value of additional knowledge acquisition does not come without cost. In our simple example, this cost has not been salient. Imagine, however, a complex decision problem in which tens or hundreds of probabilities and utilities need to be assessed. Refining each assessment in this case would almost certainly overwhelm any decision maker.

In this section, we develop a method in which decision theory is used at the metalevel to balance the benefits of extending the conversation in a given region of a model against the costs associated with extension. In doing so, we provide a tool that can help to direct a knowledge engineer or expert to areas of a model that are most deserving of attention during knowledge acquisition.

To begin, we focus on the inputs required in our approach. For each assessment that it not made with complete confidence, our approach dictates that two pieces of information be specified: a measure of the degree to which the assessment might change with additional modeling, and a measure of the cost associated with this task. We examine each of these measures in turn.

Consider an event or outcome X for which the decision maker is unwilling or unable to provide a precise probability or utility. First, the decision maker is asked to imagine a specific *assessment scenario* for X. For example, he might imagine simply extending the conversation, or he might imagine both extending the conversation and observing one or more conditioning events before a decision is made. The exact nature of the scenario is determined by the decision maker. Then, to measure the degree to which an assessment might change, the decision maker is asked to a specify a probability distribution over the variable $\phi(X)$, defined to be the value of the probability of X or utility of X *after* the assessment scenario is complete. We have already seen an idealized example of such a distribution in Figure 1.

To be practical, our approach does not require a precise distribution over $\phi$. The precise specification of such a distribution probably would take more time than a full extension of the conversation for X. Instead, only an estimate of the distribution is needed. For example, the 25 and 75-percent fractiles of the distribution provided by the decision maker could be fitted to a convenient parametric curve, such as the *beta* distribution. Alternatively, the decision maker might be allowed to draw a rough sketch of the distribution.

A measure of the cost associated with the assessment scenario imagined for X also is required in our approach. The decision maker is asked to specify this quantity directly. Of course, this cost might be uncertain. For example, the decision maker might imagine an assessment scenario of extending the conversation followed by the implementation of as-yet-undetermined experiments. In such situations, however, we require only an approximate mean. As before, forcing the expert to be completely precise would result in modeling costs that outweigh a full extension of the conversation for X.

Now let us consider how these measures can be used to balance the cost and benefits of additional assessment. Imagine that a working model of a decision problem has been developed and that a decision maker suspects that the set of events[4] and outcomes $\{X_1, X_2, ... X_n\}$ deserves attention. First, in our approach, the value of additional modeling is calculated by computing the value of perfect information on the set of variables $\{\phi(X_1), \phi(X_2), ... \phi(X_3)\}$. A Monte Carlo technique is useful for such a computation [10]. Next, the cost of additional modeling is calculated by computing the *sum* of the costs assigned to each of these variables for the given scenario.[5] The assessment scenario is recommended if and only if the computed value exceeds the computed cost.

This approach helps to focus the attention of the decision maker during model construction. If the benefits of applying a particular assessment scenario to a set of variables exceeds the costs, our method assures the decision maker that the scenario is worthwhile to pursue. If the costs outweigh the benefits, the approach leads to decision maker to consider other scenarios and other combinations of variables. In either case, the method can be used many times during knowledge acquisition to guide the decision maker to areas of the model that deserve attention.

To illustrate the approach, let us return to the football example discussed earlier. Suppose the decision maker has established the decision model shown in Figure 2 and that he has approximately 1 hour to make the bet before his gambling associate retracts the offer. Our decision maker is not comfortable with his assessment of p(Win) and wonders whether it is worth his time to extend the conversation on this event. To evaluate the tradeoff, he specifies a rough distribution for $\phi$(Win), defined to be his probability for a win after extending the conversation. The distribution is constructed by fitting a beta distribution to his estimate that the 25 and 75-percent fractiles are 0.5 and 0.6, respectively. In addition, he determines that the cost associated with 1 hour of analysis is roughly $50. In this case, a Monte Carlo analysis yields a value of perfect information of approximately $150, so it is recommended that our decision maker take the time to extend the conversation.

---

[4]The events in the set may be conditional or unconditional.

[5]In the computations of value and cost, we assume that the assessments of the $\phi(X_i)$ are independent. This assumption will be discussed in a later section.

This simple example is given to illustrate the approach. It is unlikely, in practice, that the procedure would be used on a model of such small proportions. In such cases, the decision of whether or not to refine the model usually will be intuitively obvious. Instead, the method is likely to become attractive when the number of variables analyzed is large and the resolution of assessment tradeoffs is no longer obvious.

We conclude this section with several comments about the definition of $\phi(X)$. First, we note an important restriction on the assessment scenario implicit in the definition of $\phi(X)$. In principle, a decision maker can imagine any assessment scenario. For example, a decision maker could imagine a scenario in which infinite thought is applied to assessment of X. Alternatively, he could imagine a scenario in which all propositions relevant to X, except X itself, are observed. Such scenarios might be useful in constructing models of confidence. However, for purposes of decision making, an assessment scenario should be limited to efforts that take place *before* a decision is made. There is no value to an assessment scenario in which probabilities or utilities change after the decision is made. Beyond this restriction, the scenario can take any form that is advantageous to the decision maker. In the example we discussed, the decision maker knows that he has only 1 hour to make a decision, and thus the assessment scenario is taken to be an extension of the conversation during this time. Alternatively, the decision maker may know that he can make the bet as late as the start of the game. In this case, the assessment scenario could be defined as both an extension of the conversation and the observation of conditioning events.

Second, note that the probability distribution over $\phi(X)$ is closely related to the notion of confidence. Wide distributions correspond to probability and utility assessments that are held with little confidence whereas sharply peaked distributions correspond to assessments held with a high degree of confidence. We do not claim, however, that a probability distribution over $\phi(X)$ is an accurate *descriptive* representation of confidence. An argument against such a claim is that it does not seem likely that a person imagines precisely-defined assessment scenarios when introspecting directly about confidence. Perhaps, however, a vague class or ill-defined classes of scenarios are being implicitly imagined when an individual is asked to specify a "degree of confidence" for a given assessment. It would be interesting to characterize such classes, if they exist, through empirical studies.

## Relationship to other approaches

In this section we discuss the relationship of our approach to existing methodologies for representing confidence and to classic methods for focusing attention in model construction. We also examine the relative merits of each approach.

A traditional method for focusing attention during model development is *stochastic sensitivity analysis* [9]. In this approach, a probability or utility is swept through its full range of possible values, and the expected utility for each alternative is plotted as a function of the value of the variable. The plot reveals the sensitivity of a decision to the variable being analyzed.

A stochastic sensitivity analysis for p(Win) in the simple football example is shown in Figure 3. In this case, a small downward change in p(Win) near the baseline value of 0.53 results in a reversal of the optimal alternative. Moreover, the difference in expected utilities between the two alternatives, Bet and Do not bet, is a relatively strong function of p(Win). Because our decision maker was not completely confident about his assessment of p(Win), it seems prudent in this case to allocate additional resources to assessment.

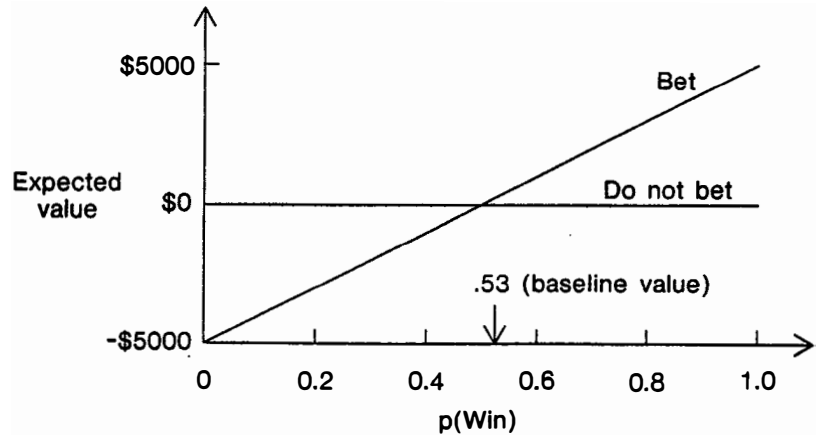

Figure 3: Sensitivity Analysis for p(Win).

Stochastic sensitivity analysis has proved to be a valuable tool for focusing attention in model development [9]. Indeed, we do not suggest that our approach replace this tool. We do suggest, however, that our approach be used when examining the sensitivity of a model to multiple variables simultaneously. In effect, our approach automates the decision process that a person goes through when confronted with a sensitivity analysis. This automation is probably overkill for the analysis of a single variable, but it is potentially useful for the analysis of multiple variables, a situation in which a traditional sensitivity analysis is difficult to display and to comprehend.

A method for representing confidence that has received renewed attention in the literature involves the assignment of *bounds* or *intervals* to the probability of events and to the utility of outcomes [11, 12, 13]. In this approach, it is assumed that a point value exists for each variable, but that the value is known only to lie in a certain range. Given a set of variables with assigned intervals, bounds on a target variable can be computed by sweeping each variable in the set through its possible range and, for every combination, inferring a point-value for the target variable using the laws of ordinary probability.[6] For example, suppose we are given the assignments

$$p(A) = [0.4, 0.6] \qquad p(B|A) = [0.3, 0.8]$$

Using this method for propagating bounds, it follows that

$$p_{max}(AB) = p_{max}(A)\, p_{max}(B|A) = (0.6)(0.8) = 0.48$$

$$p_{min}(AB) = p_{min}(A)\, p_{min}(B|A) = (0.4)(0.3) = 0.12$$

In other words, the value of p(AB) lies in the interval [0.12, 0.48].

This method of propagating bounds provides another approach for automating sensitivity analysis. Indeed, the interpretation underlying this method has been referred to as the "Bayesian sensitivity analysis interpretation" [14]. One advantage of the bounds approach over our focusing method is that it requires only two assessments for each variable; our method requires the assessment of a probability distribution and a measure of the cost of assessment. A disadvantage of the bounds method over our approach, however, is that the reduced information content makes it difficult to use the results of the analysis to determine whether a variable or set of variables should be modeled further. In particular, the results of interval analysis do not convey a clear picture of the *degree* to which the model is sensitive to a variable or set of variables. An accurate reflection of degree of sensitivity is useful for balancing the benefits of further modeling with the associated costs. Also, as we mentioned, we can reduce the information requirement of our approach by fitting a parametric distribution to a small number of assessments. It is likely that the advantages of having a full distribution will outweigh the disadvantages associated with forcing a fit that might not be completely accurate.

---

[6]The Dempster-Shafer theory of belief functions also has an interpretation in terms of bounds, but the method of propagation is believed to be incommensurate with the method described here [14].

Finally, we mention that Critchfield has developed a method for automating sensitivity analysis that uses "second order" probabilities and utilities [10]. One difference between his approach and ours is that we have tried to provide a precise interpretation for the $\phi$ distributions that is relevant to allocation of resources during knowledge acquisition. Critchfield simply proposes that distributions be assessed by requesting a person's "probability of his probability." Another difference is that Critchfield uses heuristic methods to evaluate the sensitivity of variables; these methods are limited to decisions having only two possible alternatives. Our approach, based on decision theory, is not similarly limited.

## Future work

In this section, we discuss questions raised by our approach and suggest areas for future research.

Our approach has assumed that the rough measures of confidence and cost supplied by a decision maker will produce meaningful recommendations for focusing attention. There are two questions surrounding this assumption. First, to what degree are these measures *precise* or repeatable? Second, to what degree are these measures *accurate*? That is, do the measures of confidence and cost truly reflect the beliefs and preferences of the decision maker? Both of these questions can be answered empirically. The precision of these assessments can be evaluated by giving a decision maker identical assessment tasks, separated by distractions, and by comparing the results. The accuracy of these assessments can be measured by determining whether recommendations made by our approach agree with the intuition of a decision maker in simple decision models.

There are costs associated with the assessments required by our focusing technique. Do the benefits provided by the approach outweigh these costs? Most probably, they do in some cases and do not in others. For example, we mentioned that our approach is unlikely to have a net benefit for simple decision models whereas it may have a net benefit for complex models. Are there other aspects of decision models that make them more or less suitable for our approach? Can a decision maker recognize, in a straightforward manner, situations in which the focusing method is appropriate? Empirical studies are needed to answer these questions.

The cost and benefits of the focusing technique itself can be modulated. For example, we have assumed independence among the $\phi$ distributions and among the cost measures used by the approach. This assumption can be relaxed by explicitly representing dependencies among the assessments, providing additional benefit at additional cost. As another example, whenever our technique fails to identify a sensitive cluster of variables, a person has the option to consider other assessment scenarios for the same cluster, consider other clusters, or terminate the analysis. Various costs and benefits are associated with each of these choices. Can such costs and benefits be balanced intuitively? If so, our approach is potentially useful. If not, our approach is merely a first step in a useless infinite regress. Again, empirical studies are needed to answer this important question.

Finally, we have suggested that the task of identifying clusters of variables for analysis by our approach be delegated to the decision maker. However, sensitive combinations might easily be missed. Are there useful heuristic methods for automating this search? Perhaps an algorithm guided by the qualitative structure of the network would be effective. In any case, artificial intelligence techniques are likely to be useful here.

## Summary

We examined the Bayesian paradigm for accommodating beliefs and preferences that are held with partial confidence. An important notion highlighted by the method is that additional modeling can be valuable when assessments are not held with complete confidence. We next developed a method that uses decision theory to balance the benefits of additional modeling with the associated costs. We saw that the approach can help to focus the attention of a knowledge engineer or expert on parts of a decision model that deserve additional refinement. It is hoped that our approach will facilitate the construction of more robust decision models, and help make the construction of such models more efficient.

## Acknowledgements

We thank Greg Cooper, I.J. Good, Eric Horvitz, Amos Tversky, and Mike Wellman for useful discussions. Support for this work was provided by the National Library of Medicine under grant R01-LM04529, NASA-Ames, the Henry J. Kaiser Family Foundation, and the Ford Aerospace Corporation. Computing facilities were provided by the SUMEX-AIM resource under NIH grant RR-00785.